\documentclass[runningheads]{llncs}
\usepackage{graphicx}
\usepackage{algorithm}
\usepackage{algpseudocode}

\usepackage{amsmath,amssymb}
\usepackage{multirow}
\usepackage[small]{subfigure}
\usepackage{color}
\usepackage{caption}
\usepackage[flushleft]{threeparttable}

\usepackage{tablefootnote}

\begin{document}

\title{REPrune: Filter Pruning via Representative Election} 


%
\author{Mincheol Park \and
Woojeong Kim \and
Suhyun Kim}
\authorrunning{M. Park, W. Kim and S. Kim}
%
\institute{Korea Institute of Science and Technology \\ Seoul, South Korea \\
\email{\{mincheol.park,woojeongkim,suhyun\_kim\}@kist.re.kr}}
\maketitle

\begin{abstract}
Even though norm-based filter pruning methods are widely accepted, it is questionable whether the ``smaller-norm-less-important" criterion is optimal in determining filters to prune. Especially when we can keep only a small fraction of the original filters, it is more crucial to choose the filters that can best represent the whole filters regardless of norm values. Our novel pruning method entitled ``REPrune" addresses this problem by selecting representative filters via clustering. By selecting one filter from a cluster of similar filters and avoiding selecting adjacent large filters, REPrune can achieve a better compression rate with similar accuracy. Our method also recovers the accuracy more rapidly and requires a smaller shift of filters during fine-tuning. Empirically, REPrune reduces more than 49\% FLOPs, with 0.53\% accuracy gain on ResNet-110 for CIFAR-10. Also, REPrune reduces more than 41.8\% FLOPs with 1.67\% Top-1 validation loss on ResNet-18 for ImageNet. 

\end{abstract}

\section{Introduction}
The increase in the number of convolution layers is a crucial cumbersome to deploy deep convolutional neural network (CNN) models. According to the recent study~\cite{cong2014minimizing}, convolution operations account for over 90\% of the total volume. Novel architectural concepts reshaped a model into a computation efficient network by replacing spatial convolutions with a depthwise manner \cite{chollet2017xception}, \cite{howard2017mobilenets}, \cite{sandler2018mobilenetv2}. However, these approaches still suffer from frequent memory access to supply parameters for multiply-add computations \cite{chen2016eyeriss}, \cite{yang2017designing}.

\begin{figure}[!t]
\centering{
\subfigure[REPrune via filter clustering] {\includegraphics[width=0.48\linewidth]{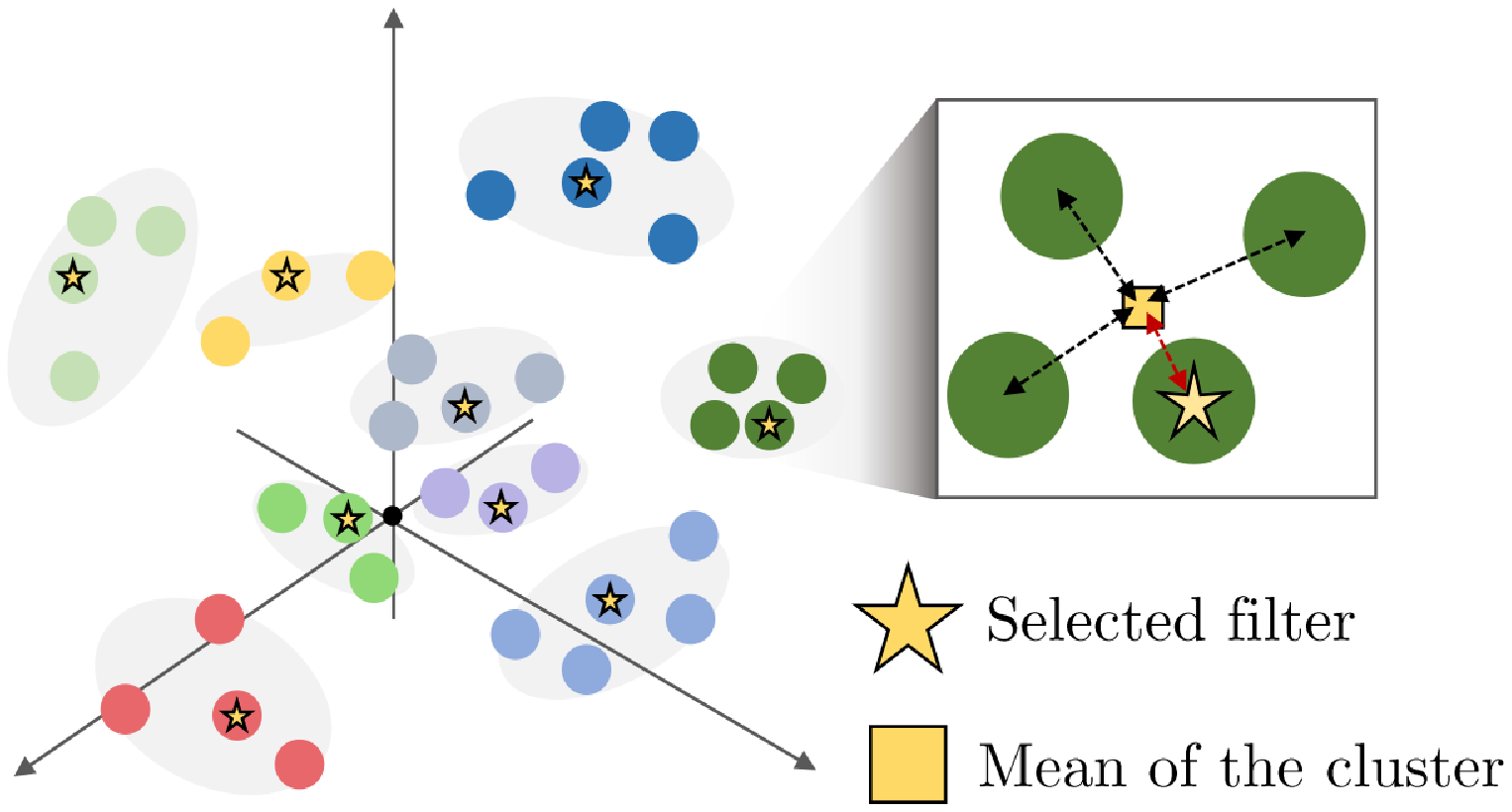}
\label{fig1_a}}
\subfigure[Smaller norm less importance]{\includegraphics[width=0.48\linewidth]{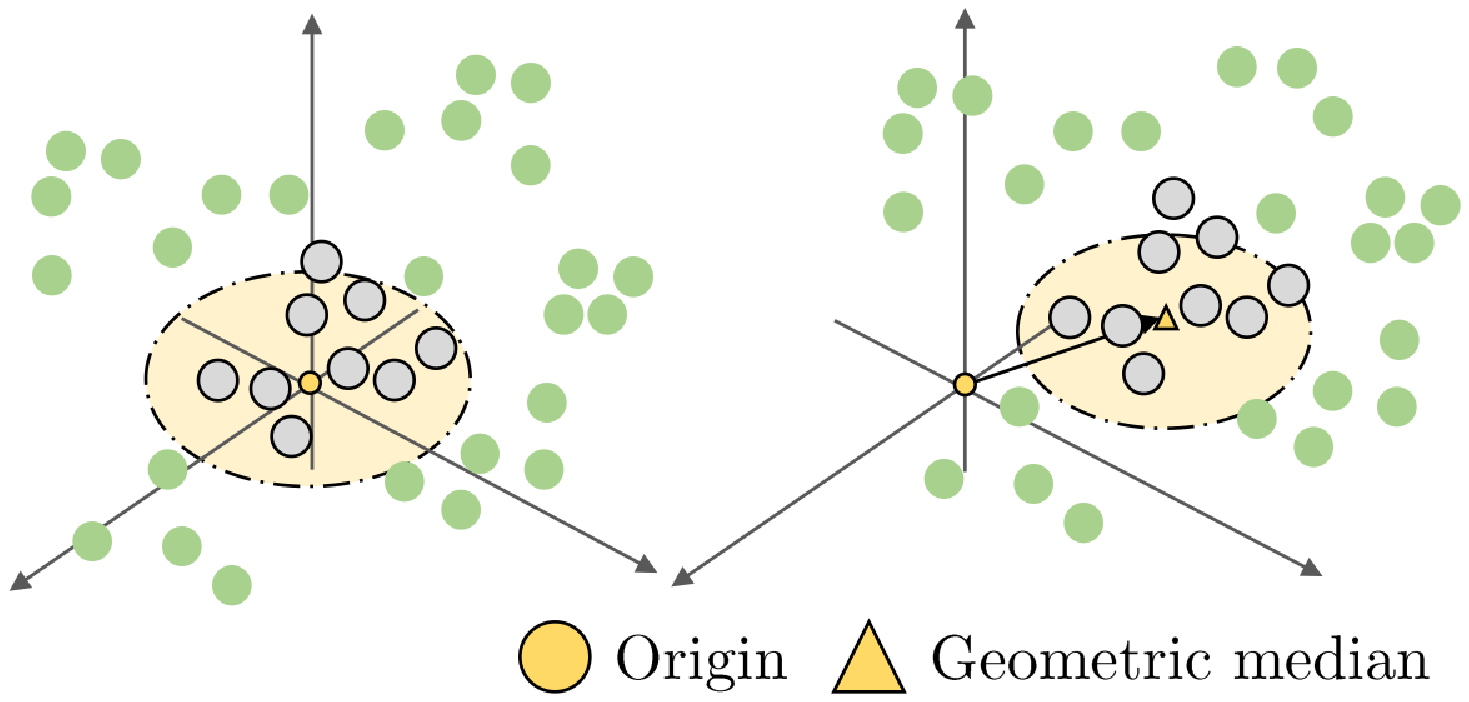}
\label{fig1_b}}
}
\caption{Our pruning method compared to the well-known criteria: {\bf (a)} illustrates our approach, REPrune, which selects a representative filter from each cluster. Different colors of dots indicate they are in different clusters. The filter closest to the mean of the cluster is elected as the representative (denoted by a star mark). {\bf (b)} illustrates norm-based pruning criteria from the central point. Light green dots denote the remaining filters, and gray dots in the yellow area denote the pruned filters. The central point is the origin and geometric median, respectively.}
\end{figure}

Prior works have tried to reduce the redundant parameters to achieve a small-sized model suitable for real-world scenarios. For example, the weight pruning \cite{han2015deep} iteratively removed connections considered not to be important. But, the induced unstructured sparse matrices could weaken the effects of leveraging Basic Linear Algebra Subprograms (BLAS) libraries. On the contrary, the filter pruning \cite{he2017channel}, \cite{liu2017learning} discarded filters in convolution layers based on importance strategy. As these methods produce sparse channel models with structured variables, they could benefit from BLAS acceleration with the memory and computation gain.

Nevertheless, the previous filter pruning methods are still questionable in the point of view: {\bf whether smaller norm from a centered point truly indicates less importance}. As shown in Fig.~\ref{fig1_b}, the most recent studies \cite{li2016pruning}, \cite{he2018soft}, \cite{he2019filter} conduct the filter pruning based on ``norm criterion," which regards the filters with smaller distance, e.g.,$\ell_1$ or $\ell_2$, from the origin or geometric median as trivia. However, if an extremely high pruning ratio is applied to meet strongly lightweight requirements, e.g., tens or hundreds of MFLOPs~\cite{zhang2018shufflenet}, for actual deployment~\cite{alyamkin2019low}, the model remains severely damaged on accuracy due to relying on the filters existing only on the outlier. \cite{ye2018rethinking} somewhat verified enlarging norm appears highly informative. We empirically found norm alone cannot detect highly informative filters, which have trouble to success classification in hugely motivated lightweight CNNs.

Our motivation starts with the consideration regarding filter distribution. As convolution filters gradually unfold the entangled manifold of features, some of the filters are locally clustered \cite{denil2013predicting}, \cite{denton2014exploiting}. However, norm-based pruning criteria consider only absolute distance, not the relation between filters. We aim to exploit relative distance for the filters. We propose ``REPrune" to choose necessary representative filters in relatively clustered them, as shown in Fig.\ref{fig1_a}. REPrune makes clusters with agglomerative clustering, regarding their distribution in each layer and evaluate which clusters may later be effective through Silhouette Coefficient~\cite{rousseeuw1987silhouettes}. Selected representatives in the clusters play a role as a locally informative key, including the similarity of features. That is, REPrune is no longer limited to pushing into high compression pressure, unlike norm-based methods did. For example, REPrune shows that ResNet-56, which is known for the optimal capacity model, presents at most 2.6\% accuracy drops despite 8.84x acceleration with 83.3\% total reduced parameters on CIFAR-100.

Our contributions are as follows: 
(1) REPrune proposes new pruning criteria based on the distribution of filters, not the magnitude of filters.
By reflecting the distribution, the selected filters are more informative in terms of how many original filters can be approximated. 
(2) Our method copes with demanding pruning requirements better. 
It selects only one filter from a cluster of neighboring filters avoiding selecting coincident large filters. This enables REPrune to perform effectively even when pruning highly optimized models such as ResNet. 
(3) The filters selected by REPrune are superior as they are,
which is supported by the rapid accuracy recovery during fine-tuning, and the higher cosine similarity between the filters before and after fine-tuning.

\section{Related Works} \label{related_works}
Previous CNN acceleration studies can be split into five categories: tensor factorization~\cite{jaderberg2014speeding}, low precision quantization~\cite{wang2018training}, architectural search~\cite{dong2019network}, knowledge distillation~\cite{park2019relational}, and pruning. The pruning approaches pursue removing unnecessary weights in charge of connections in CNNs. To the best of our knowledge, pruning methods divided into three aspects, namely, weight pruning, channel pruning, and filter pruning.


{\bf Weight pruning} focuses on removing the unimportant connections, which have less sensitivity to accuracy. For example, \cite{han2015learning} iteratively eliminates small weights less than heuristically predefined values. \cite{yang2017designing} proposes to prune the small weights according to the priority of highly energy consumed layers. However, these kinds of works incur unstructured weight matrices such as the compressed sparse row (CSR) or the coordinate list (COO) \cite{han2015deep}, which require special hardware to accelerate deployment with this transformed layout \cite{han2016eie}.

{\bf Channel pruning} is well known in terms of removing nodes along with filter pruning. Most studies grant data-driven sparsity to the coefficient factors corresponding to the channels. \cite{liu2017learning} imposes sparse constraints on the scaling factors in the batch normalization during scratched training. \cite{he2017channel} obtains the optimal network configuration by a LASSO regression-based channel selection. \cite{you2019gate} gradually removes less important channels at the training by prioritizing channels based on combination gradients by data loss with scaling factors in batch normalization. \cite{yu2017scalpel} proposes a cascading trainable mask after the convolution layers to determine which channels turn on or off at the inference. Since those methods force intended sparsity loss to accompany data loss, they struggle hard to avoid natural performance drops on the model.

{\bf Filter pruning} is also one of the node pruning methods, and it determines which filters to be evicted based on the magnitude of filters. \cite{li2016pruning} discards the filters with smaller $\ell_1$ norm. \cite{he2018soft} removes the filter in the same way as \cite{li2016pruning}, but differs in that they uses $\ell_2$ norm. They also propose soft fine-tuning, which maintains the model capacity. \cite{he2019filter} discovers that filter distribution varies in each layer, which proposes shifting the origin to the geometric median of filters for the norm calculation. However, these approaches have trouble in selecting distinctively informative filters because they consider importance only by a distance from a single point without reflecting the relative position of filters. 

Our work belongs to this filter pruning category. We select one filter from the cluster of correlated filters, which is different from absolute distance criteria.  That is, our approach implicitly has properties of Low-Rank Approximation. However, our method differs from their methods, such as algebraic decomposition~\cite{jaderberg2014speeding} or additional loss as a forced regularization~\cite{wen2017coordinating}.


\section{Methodology} \label{methodology}
We propose a novel filter pruning method entitled REPrune, which takes the distribution of filters into consideration. We cluster the filters of each convolution layer using the agglomerative clustering and pick the best one from each cluster. Our proposed method consists of three parts: agglomerative clustering, finding optimal cluster number with Silhouette Coefficient, and selecting representative filters.

\subsection{Preliminaries}
Assume that there are $L$ layers in a model. Let $n_i$ and $n_{i+1}$ denote the number of input channels and the output channels of the $i$th convolution layer, and $h_i$ and $w_i$ denote the height and the width of the filters in the $i$th layer. The filter weight of the $i$th layer $\mathbf{W}^{(i)} \in \mathbb{R}^{n_{i+1} \times n_{i}\times h_{i}\times w_{i}}$ consists of $n_{(i+1)}$ filters. The $j$th filter of the $i$th convolution layer is denoted as $\mathcal{F}_{i,j} \in  \mathbb{R}^{n_{i}\times h_{i}\times w_{i}}$. 

\subsection{Agglomerative Clustering}
Agglomerative clustering recursively merges the pair of clusters by minimally increasing a given linkage distance. Assume $K_i$ is the number of clusters for the $i$th convolution layer. Agglomerative clustering for the $i$th convolution layer filter is performed as follows.  

\begin{quotation}
\noindent
1. Each filter $\mathcal{F}_{i,j}$ is considered as an individual cluster. \\
2. For all the clusters, compute the distance to all other clusters. \\
3. Combine two clusters with the smallest distance into a cluster. Therefore, the number of clusters is reduced by one. \\
4. Repeat 2 and 3 until $K_i$ clusters are formed.
\end{quotation}

In step 3, the Ward's method~\cite{ward1963hierarchical} is used to measure the distance between two clusters, which minimizes the variance of the clusters being merged. The distance between two filter clusters, $\mathcal{C}_A$ and $\mathcal{C}_B$, is defined as how much the sum of squares will increase when two clusters are merged:
\begin{equation}\label{ward}
\begin{split}
d(\mathcal{C}_A, \mathcal{C}_B) & =\sum_{\mathcal{F}_{\cdot,j}\in \mathcal{C}_A\cup \mathcal{C}_B}{\left \| {\mathcal{F}_{\cdot,j}} - \mathbf{m}_{\mathcal{C}_A \cup \mathcal{C}_B} \right \|}_2 \\
                                      & - \sum_{\mathcal{F}_{\cdot,j}\in \mathcal{C}_A}{\left \| {\mathcal{F}_{\cdot,j}} - \mathbf{m}_{\mathcal{C}_A} \right \|}_{2} - \sum_{\mathcal{F}_{\cdot,j}\in \mathcal{C}_B}{\left \| {\mathcal{F}_{\cdot,j}} - \mathbf{m}_{\mathcal{C}_B} \right \|}_{2} \\
                                      & = \frac{\left | \mathcal{C}_A \right | \left | \mathcal{C}_B \right |}{\left | \mathcal{C}_A \right | + \left |  \mathcal{C}_B \right |} {\left \| \mathbf{m}_{\mathcal{C}_A} - \mathbf{m}_{\mathcal{C}_B} \right \|}^2
\end{split}
\end{equation}
, where $\mathbf{m}_{\mathcal{C}_A}$ is the centroid of cluster $\mathcal{C}_A$, and $\left |  \mathcal{C}_A \right |$ is the number of filters in it. Therefore, the sum of squares starts out at zero when every filter is in its own cluster and then grows as we merge clusters in a hierarchical manner. Ward’s method keeps this growth as small as possible.

We choose agglomerative clustering for two reasons: First, agglomerative clustering can take more account into locally grouped filters in filter distribution because it recursively merges the nearest pair. Second, unlike centroid-based clustering such as K-means clustering, it is not affected by random initialization of centroids, stabilizing clustering performance. 

\subsection{Finding the Optimal Number of Clusters} \label{optimal}
Most of the previous filter pruning works applied the same static pruning ratio to all layers. However, our proposed method dynamically determines the number of clusters appropriate for the filter distribution. We employed Silhouette Coefficient \cite{rousseeuw1987silhouettes} to find the optimal number of clusters.

For each filter $\mathcal{F}_{\cdot, j}$ in the cluster $\mathcal{C}_k$, let $a(\mathcal{F}_{\cdot, j})$ be the mean distance between $\mathcal{F}_{\cdot, j}$ and all other filters $\mathcal{F}_{\cdot, j'}$ in the same cluster:
\begin{equation}\label{silhouette_a}
a(\mathcal{F}_{\cdot, j}) = \frac{1}{\left | \mathcal{C}_k \right |-1} \sum_{\mathcal{F}_{\cdot, {j}'} \in C_k, j\neq j'}{\left \| \mathcal{F}_{\cdot, j} - \mathcal{F}_{\cdot, j'} \right \|_2}
\end{equation}
$a(\mathcal{F}_{\cdot, j})$ measures the degree of cohesion, how dense the filters are in the cluster. Hence the smaller the value of $a(\mathcal{F}_{\cdot, j})$, the better the assignment.

Let $b(\mathcal{F}_{\cdot, j})$ be the smallest mean distance of $\mathcal{F}_{\cdot, j}$ to all filters in any other cluster $\mathcal{C}_{k'}$, of which $\mathcal{F}_{\cdot, j}$ is not included:
\begin{equation}\label{silhouette_b}
b(\mathcal{F}_{\cdot, j}) = \min_{k \neq k'}\frac{1}{\left | \mathcal{C}_{k'} \right |} \sum_{\mathcal{F}_{\cdot, j'} \in \mathcal{C}_{k'}}{\left \| \mathcal{F}_{\cdot, j} - \mathcal{F}_{\cdot, j'} \right \|_2}
\end{equation}
We can interpret $b(\mathcal{F}_{\cdot, j})$ as how close each filter is to filters in the neighboring cluster, which measures the degree of separation. It is considered more desirable to assign different clusters farther, so the larger the value of $\mathcal{F}_{\cdot, j}$ is, the better.

Now, the Silhouette Coefficient for a single filter $\mathcal{F}_{\cdot, j}$ is defined as:
\begin{equation} \label{silhouette_filter}
    s(\mathcal{F}_{\cdot, j}) = 
    \begin{cases}
    \displaystyle \frac{b(\mathcal{F}_{\cdot, j}) - a(\mathcal{F}_{\cdot, j})}{\max\{a(\mathcal{F}_{\cdot, j}), b(\mathcal{F}_{\cdot, j})\}},& \text{if } {\left | \mathcal{C}_k \right |} > 1\\
    0,              & \text{if } {\left | \mathcal{C}_k \right |} = 1
    \end{cases}
\end{equation}
The Silhouette Coefficient ranges from -1 to +1. Filters with high Silhouette value are close to filters in the same cluster and far from the filters in other clusters. However, filters with a small or negative $s$ are likely to be placed in the wrong cluster. 

\begin{figure*}[!t]
\centering{
\subfigure[4th and 5th convolution layer of VGG-16 on CIFAR-10] {\includegraphics[width=0.98 \columnwidth]{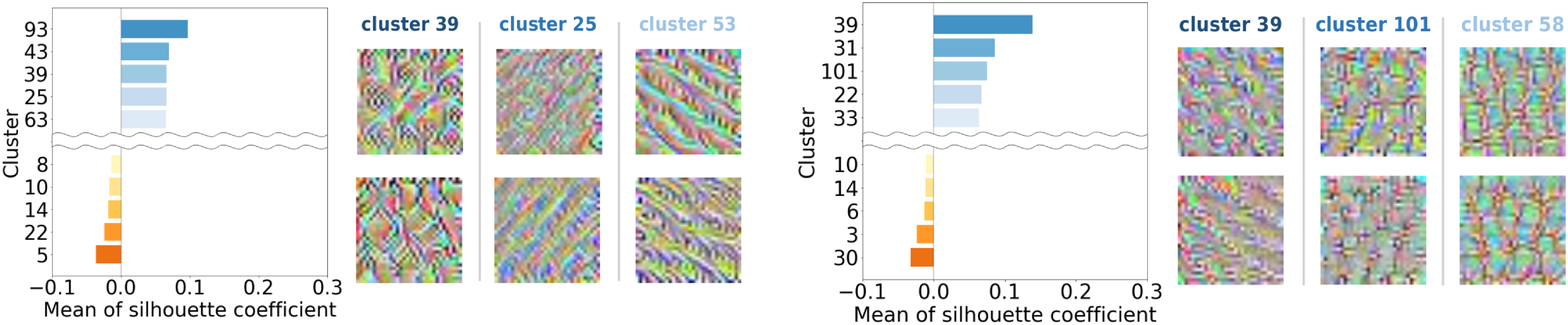}
\label{fig2_a}}
\subfigure[1st and 6th convolution layer of VGG-16 on ImageNet] {\includegraphics[width=0.98 \columnwidth]{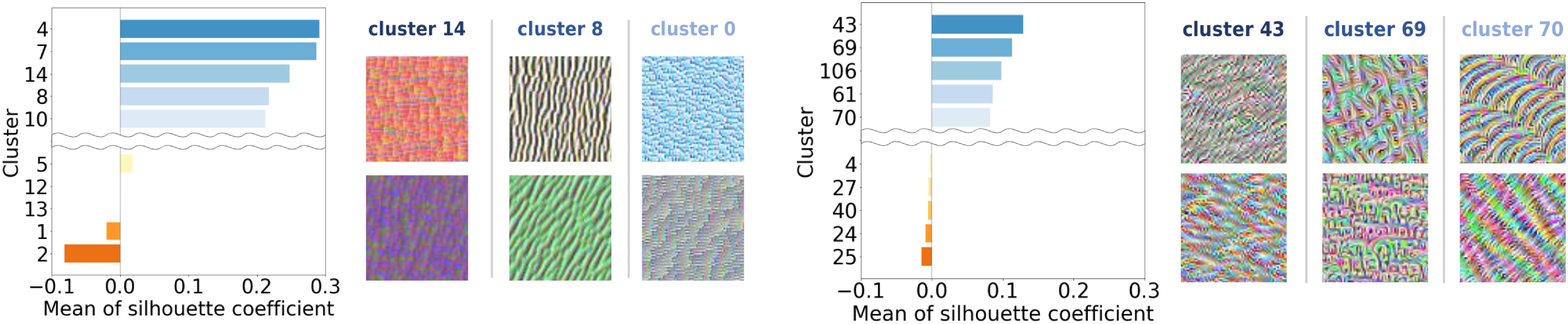}
\label{fig2_b}}
}
\caption{Silhouette Coefficient plots and visualization \cite{erhan2009visualizing} of the convolution filters in clusters with top Silhouette Coefficient after agglomerative clustering. Each row is the result of VGG-16 on CIFAR-10 and ImageNet, respectively. If the mean of the Silhouette Coefficient is higher, it indicates that the cluster is well cohesive and separated from the other ones. In convolution filter visualization, filters from clusters with high Silhouette Coefficient mean share common traits.}
\label{fig2}
\end{figure*}

By taking the average of each filter's Silhouette value, we can evaluate how well each cluster or the whole layer is aggregated. We denote $s(\mathcal{C}_k)$ as the mean Silhouette value of the filters within the cluster $\mathcal{C}_k$:
\begin{equation} \label{silhouette_cluster}
s(\mathcal{C}_k) = \frac{\sum_{\mathcal{F}_{\cdot, j} \in \mathcal{C}_k}s(\mathcal{F}_{\cdot, j})}{\left | \mathcal{C}_{k} \right |}
\end{equation}
, where $K$ is the cluster number. Higher values of $s(\mathcal{C}_k)$ indicate that the cluster $\mathcal{C}_k$ is well formed.

The Silhouette Coefficient for the whole layer is also given as the mean of the one for each filter. Silhouette Coefficient mean is used to determine the optimal number of clusters within the $i$th convolution layer:
\begin{equation} \label{optimal_cluster_num}
K^{*}_{i} = \underset{K}{\operatorname{arg\,max}}\frac{\sum_{j=1}^{n_{i+1}}s(\mathcal{F}_{i, j})}{n_{i+1}}.
\end{equation}
We set $K^{*}_{i}$ to be the optimal number of clusters and use the established clusters to select the representative filter.

\subsection{Selecting the Representative Filter}
Assuming filters in the same cluster share similar properties, we select only one representative from the cluster and prune the rest. As Fig.~\ref{fig2} shows, filters in the cluster with high Silhouette Coefficient resemble each other, justifying our selection algorithm.

As mentioned in Section~\ref{optimal}, clusters with higher Silhouette Coefficient values gather similar filters together. If $s(\mathcal{C}_k)$, the average of Silhouette Coefficient within cluster $k$, is below zero, we do not select any filter from the cluster. For all other clusters, we select on e filter from each that best represents the cluster.

We choose the filter most adjacent to the mean to represent the whole cluster. Cluster centroid $\mathbf{m}_{\mathcal{C}_k}$ is obtained by averaging filters in it:
\begin{equation}\label{cluster_mean}
\begin{split}
\mathbf{m}_{\mathcal{C}_k} = \frac{\sum_{j=1}^{n_{i+1}}\mathcal{F}_{i, j}}{n_{i+1}}
\end{split}
\end{equation}
From the cluster $\mathcal{C}_k$, find the filter nearest to the cluster mean:
\begin{equation}\label{nearest_filter}
\begin{split}
\mathcal{F}_{\cdot, j^*} = \underset{\mathcal{F}_{\cdot, {j}}}{\operatorname{arg\,min}} {\left \| {\mathcal{F}_{\cdot,{j}}} - \mathbf{m}_{\mathcal{C}_k} \right \|}
\end{split}
\end{equation}
$\mathcal{F}_{\cdot, j^*}$ be the selected filter in cluster $\mathcal{C}_k$, and we remove the rest to prune the layer.

\newcommand{\factorial}{\ensuremath{\mbox{\sc Factorial}}}
\begin{algorithm}[!b]
\caption{REPrune Algorithm}\label{REPrune}
\begin{algorithmic}
\Require{filter weights: $\mathbf{W}^{(i)} \in \mathbb{R}^{n_{i+1} \times n_{i}\times h_{i}\times w_{i}}, i = [1,...,L]$}
\State{{\bf Given:} minimum cluster rate: $\lambda$}
\State{$K_{min} \leftarrow \left \lfloor {n}_{i+1} \times \lambda \right \rfloor$}
\State{$K_{max} \leftarrow {n}_{i+1} - 1$}
\For{cluster number $K \in (K_{min} : K_{max})$}
    \State{Agglomerative Clustering $\mathcal{AC} \leftarrow \mathbf{W}^{(i)}, K$}
    \State{Compute $s(\mathcal{F}_{i, j})$ from Equation~\ref{silhouette_filter}}
\EndFor
\State{Find the optimal cluster number $K_{i}^{*}$ which satisfies Equation~\ref{optimal_cluster_num}}
\For{cluster $\mathcal{C}_k \in (\mathcal{C}_{1} : \mathcal{C}_{K^{*}_{i}})$}
    \If{$s(\mathcal{C}_{k}) >= 0$}
        \State{Find the filter which satisfies Equation~\ref{nearest_filter}} and append it to $\mathbf{W}^{(i)}_P$
    \EndIf{}
\EndFor
\Ensure{pruned filter weights: $\mathbf{W}^{(i)}_P \in \mathbb{R}^{n_{i+1} \times P \times h_{i}\times w_{i}}, i = [1,...,L]$}
\end{algorithmic}
\end{algorithm}

\subsection{REPrune Algorithm}
The overall REPrune procedure is as follows — first, train the original model and obtain the trained weight $\mathbf{W}$. Next, for each layer $\mathbf{W}^{(i)}$ in $\mathbf{W}, i = [1,...,L]$, prune filters as described in Algorithm~\ref{REPrune}. 
Minimum cluster rate, $\lambda$ is introduced to control the pruning ratio that REPrune targets. Because the number of remaining filters also has a significant effect on accuracy, REPrune adjusts the minimum pruning ratio with given $\lambda$. The pruning process is one-shot. And then, replace filter weights $\mathbf{W}$ with pruned filter weights $\mathbf{W}_P$, which produces the narrower model. Lastly, fine-tune the pruned model.

\section{Experimental Analysis} \label{experiments}
We evaluate REPrune on the two types of CNN models: VGG \cite{simonyan2014very} with huge capacity and ResNet \cite{he2016deep} with optimal capacity. Also, we apply our approach on various datasets with different scales: CIFAR-10, CIFAR-100 \cite{hinton2007learning}, and ILSVRC-2012 \cite{russakovsky2015imagenet} briefly called ImageNet. Both CIFAR-10 and CIFAR-100 contain 50,000 training images and 10,000 testing images. Each dataset is divided into 10 or 100 classes. Their image size is 32$\times$32. The ImageNet is one of the popular large-scale datasets, which includes 1.27 million training images and 50,000 validation images with 1,000 classes. ImageNet includes images with an average size of 469$\times$387. Preprocessing including normalization and image crop follows \cite{marcel2010torchvision} for ImageNet and \cite{he2019filter} for the others.

{\bf Training}. For CIFAR, we trained the model for 200 epochs with SGD, a momentum of 0.9, and a weight decay with 0.0005. The learning rate starts at 0.1, then decreases to one-tenth when the half and three-quarters of the total epoch pass. Original model comes from MSRA initialization \cite{he2015delving}. For ImageNet, we employ the pretrained weights \cite{marcel2010torchvision}. We use 128 mini-batches for all datasets. 

{\bf Fine-tuning}. For CIFAR, we set the same fine-tuning epochs as a scratch. We set 0.9 for the momentum, and 0.001 for the weight decay. For ImageNet, we fine-tune 160 epochs with the weight decay of 0.0001. The learning rate for all datasets starts at 0.01 with the same annealing strategies as a scratch. Batch size of 128 for CIFAR and 256 for ImageNet are used.

\subsection{Varying Minimum Cluster Rate $\lambda$} \label{optimalclusters}

\begin{figure*}[!t]
\centering{
\subfigure[] {\includegraphics[width=0.98 \columnwidth]{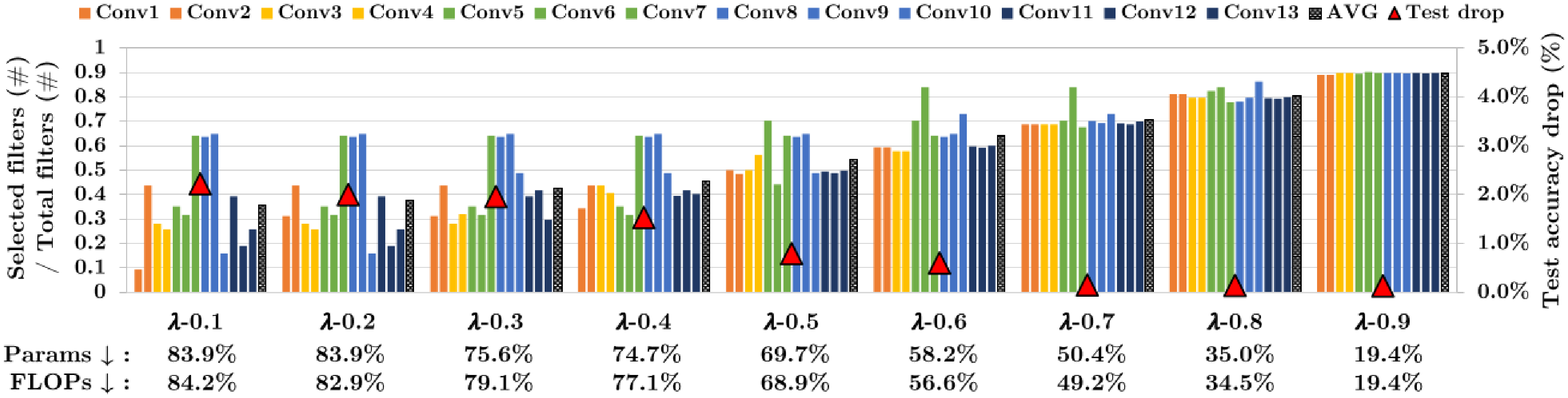}
\label{fig3_a}}
\subfigure[] {\includegraphics[width=0.98 \columnwidth]{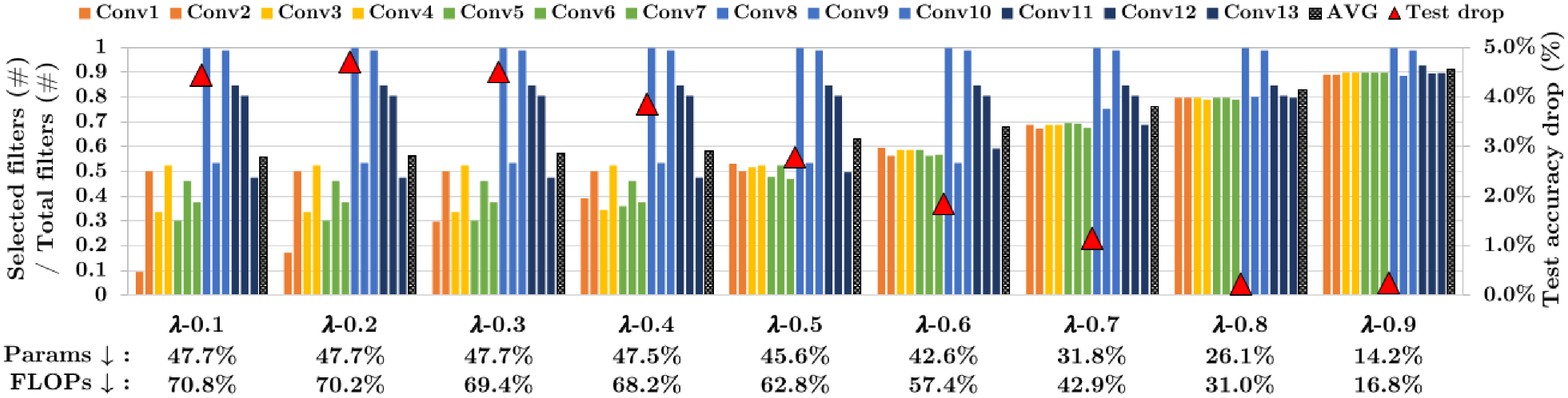}
\label{fig3_b}}
}
\caption{Remaining filter ratio of VGG-16 on both (a) CIFAR-10 and (b) CIFAR-100 with given $\lambda$. Each bar is for each layer, and the color indicates the number of original channels, such as [`Red': 64, `Yellow': 128, `Green': 256, `Blue': 512, `Dark Blue': 512]. The red triangle indicates the test accuracy drop for each $\lambda$, and $\downarrow$ denotes the reduction size.
}
\label{fig3}
\end{figure*}

In this section and Fig.\ref{fig3}, we describe how the minimum cluster rate $\lambda$ affects the number of selected filters and their relation with accuracy.

{\bf Experiments on CIFAR}. For CIFAR-10, as shown in Fig.~\ref{fig3_a}, REPrune with $\lambda$ of 0.1 to 0.3 produces an extremely compressed model with the parameter size of 75.6\% to 83.9\%. Despite the extremely reduced size of the model, the accuracy drops by up to 2.22\%. As $\lambda$ shifts to 0.9, the number of filters is strongly limited upon $\lambda$. As a result, it performs as a conventional layer-wise pruning with 0\% accuracy drop. For CIFAR-100 shown in Fig.~\ref{fig3_b}, when $\lambda$ is set from 0.1 to 0.4, the number of filters searched by REPrune remains almost the same. At the time, only the filter ratio of the first layer differs, decreasing accuracy drop to 3.8\%. Similar to CIFAR-10, the larger the lambda, the more dependent it is, and the accuracy drop converges to zero. Therefore, given low $\lambda$, REPrune has a high degree of freedom in determining the number of remaining filters. Although the accuracy drops as the absolute number of filters decreases, REPrune shows robustness in accuracy drop with the low $\lambda$ values.

{\bf Acceleration}. As described in Fig.~\ref{fig3}, REPrune produces various scales of network optimization on both CIFAR-10 and CIFAR-100. When $\lambda$ is 0.1, VGG-16 theoretically achieves 6.32x acceleration on CIFAR-10 and 3.33x on CIFAR-100. Here, the pruned network has a small accuracy drop, 1.9\%, and 4.2\% for each dataset. If $\lambda$ is 0.9, VGG-16 achieves 1.24x and 1.20x speedup for both datasets. In general, REPrune covers a wide range of acceleration regardless of data classes. Given the acceleration condition, we can fit the size of the model produced by REPrune by setting $\lambda$. 



\subsection{Adaptivity under Demanding Pruning Requirements} \label{demandingpruning}
We compare the performance of REPrune with other filter pruning methods \cite{he2018soft}, \cite{he2019filter} under severe pruning constraints.

In this experiment, we aim to measure the performance of REPrune under severe pruning constraints. We set the small value of $\lambda$, 0.1 to 0.4, and compare them with other norm-based pruning methods. SFP \cite{he2018soft}, and FPGM \cite{he2019filter} put zero masks on the filters with small $\ell_2$ distance from the origin or the geometric median, respectively. For the comprehensive analysis of norm-based methods, we experiment with $\ell_1$ norm instead of $\ell_2$ as well. The pruning ratio of four norm-based models matches that of REPrune with each $\lambda$. 
To make sure that REPrune adapts well in conditions where it can be considered highly optimized, we prune the low-parameterized model, namely, ResNet.
Unlike previous works \cite{li2016pruning}, \cite{son2018clustering}, we do not adopt constraint on sensitive low-level layers, only using the representatives obtained by REPrune algorithm.
\begin{figure*}[!t]
\centering{
\subfigure[ResNet-32] {\includegraphics[width=0.48 \columnwidth]{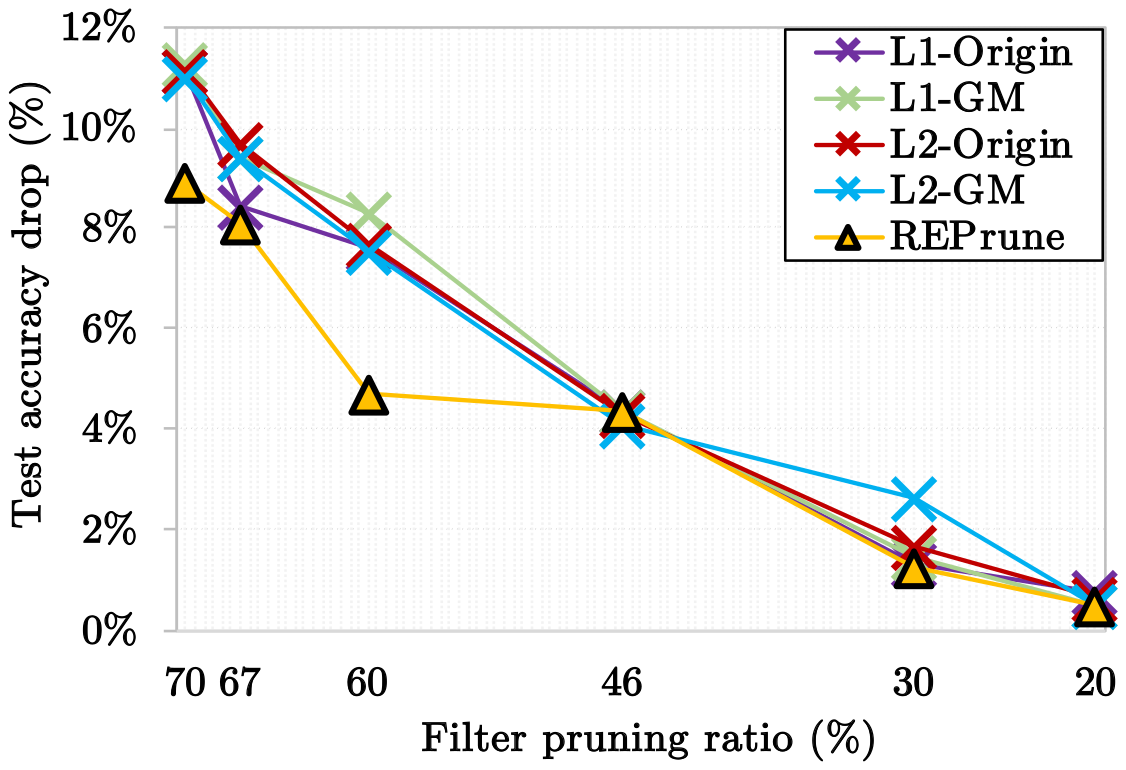}
\label{fig4_a}} 
\subfigure[ResNet-56] {\includegraphics[width=0.48 \columnwidth]{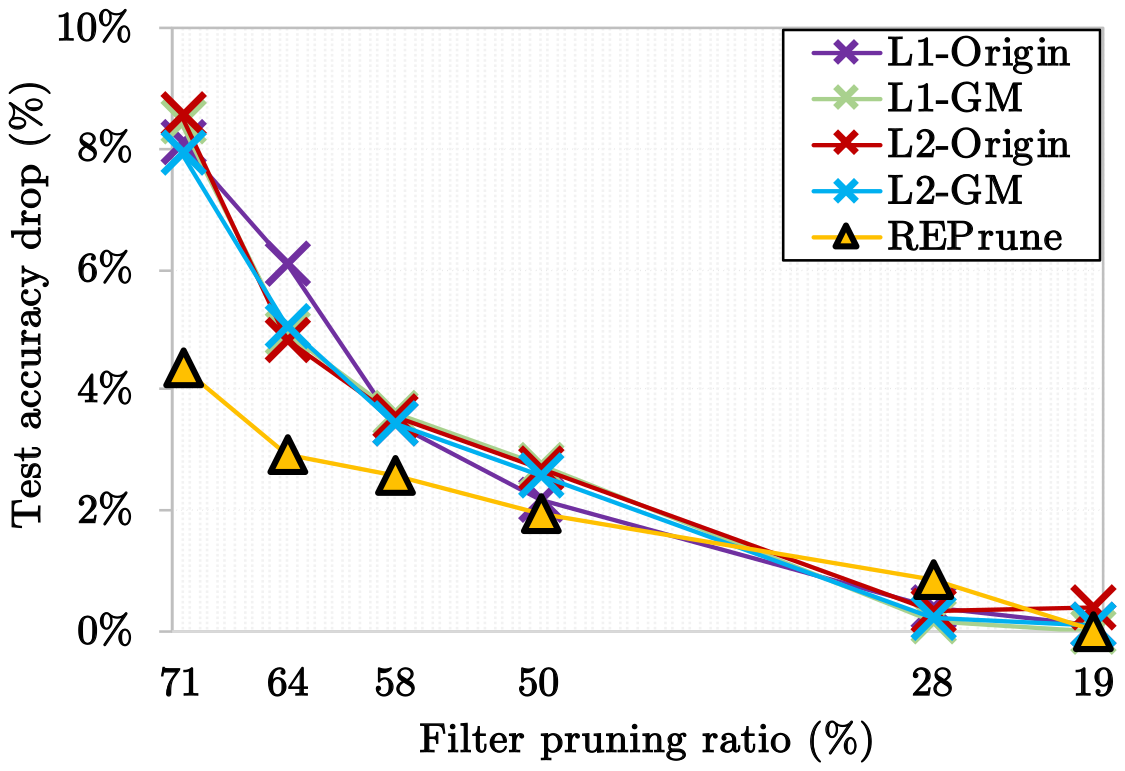}
\label{fig4_b}}
}
\caption{Comparision with $\ell_2$ norm-based filter pruning methods of ResNet-32 and ResNet-56 on CIFAR-100. Each pruning ratio corresponds to that of REPrune given $\lambda$, 0.1, 0.2, 0.3, and 0.4.
}
\label{fig4}
\end{figure*}

As shown in Fig.~\ref{fig4}, REPrune outperforms other norm-pruning methods for both ResNet-32 and ResNet-56.  For ResNet-32, the deviation of accuracy drop is trivial despite the increase in pruning ratio from 46\% to 60\%. The accuracy drop of REPrune slightly increases by 0.38\%, while that of the other methods rises by 3.51\% on average. Especially for the relatively deeper model, ResNet-56, the higher the pruning ratio, REPrune exceeds other methods by a larger margin. Thus, REPrune is more appropriate for situations that require extreme pruning constraints on ResNet models.

\subsection{Image Classification} \label{classification}
In this section, we compare our approach with various types of previous works. Here, we choose $\lambda$ to prune models with a comparable level of FLOPs against other works. For CIFAR, the best accuracy of the five experiments is chosen. 
To deal with the identity conflict in each skip-connection, point-wise convolution is exploited for down-sampling.
Furthermore, as ResNet utilizes small-sized channels, the number of channels required to fully represent the features may not be obtained after pruning. To prevent this, we use dropout for generalization.

\begin{table*}[!t]
\begin{center}
\scriptsize{
\begin{tabular}{c|c|ccc|ccc}
\hline
\multirow{2}{*}{Models} & \multicolumn{1}{c|}{\multirow{2}{*}{Methods}} & \multicolumn{3}{c|}{CIFAR-10} & \multicolumn{3}{c}{CIFAR-100} \\ \cline{3-8} 
                        & \multicolumn{1}{c|}{}                         & \multicolumn{1}{c|}{Pruned Acc.} & \multicolumn{1}{c|}{Acc. Drop} & \multicolumn{1}{c|}{FLOPs} & \multicolumn{1}{c|}{Pruned Acc.} & \multicolumn{1}{c|}{Acc. Drop} & \multicolumn{1}{c}{FLOPs} \\ \hline
\multirow{4}{*}{20}     & MIL~\cite{dong2017more}  & \textbf{91.68\%} & \textbf{1.06\%} & 2.61E7 (36.0\%) & 64.66\% & 2.87\% &  2.73E7 (33.1\%)  \\
                        & SFP~\cite{he2018soft}  & 90.83\% & 1.37\% & 2.43E7 (42.2\%) & 64.37\% & 3.25\% & 2.43E7 (42.2\%)  \\
                        & FPGM~\cite{he2019filter} & 91.09\% & 1.11\% & 2.43E7 (42.2\%) & \textbf{66.86\%} & \textbf{0.76\%} & 2.43E7 (42.2\%)  \\
                        & REPrune   & 91.28\% & \textbf{1.06\%} & \textbf{2.23E7} \textbf{(46.0\%)} & 65.81\% & 2.47\% & \textbf{2.10E7} \textbf{(49.5\%)}  \\ \cline{1-8}\cline{1-8}
\multirow{4}{*}{32}     & MIL~\cite{dong2017more}  & 90.74\% & 1.59\% & 4.76E7 (31.2\%) & 67.39\% & 2.69\% &  4.32E7 (37.5\%)  \\
                        & SFP~\cite{he2018soft}  & 92.08\% & 0.55\% & 4.03E7 (41.5\%) & 68.37\% & 1.40\% & 4.03E7 (41.5\%)  \\
                        & FPGM~\cite{he2019filter} & \textbf{92.31\%} & \textbf{0.32\%} & 4.03E7 (41.5\%) & 68.52\% & 1.25\% & 4.03E7 (41.5\%)  \\
                        & REPrune   & 92.26\% & 0.97\% & \textbf{4.08E7} \textbf{(42.0\%)} & \textbf{69.47\%} & \textbf{1.29\%} & \textbf{3.74E7} \textbf{(46.9\%)}  \\ \cline{1-8}
\multirow{7}{*}{56}     & PFEC~\cite{li2016pruning} & 91.31\% & 1.75\% & 9.09E7 (27.6\%) & $-$ & 
                        $-$ &  $-$  \\
                        & CCK~\cite{son2018clustering} & 93.24\% & 0.48\% & 8.83E7 (30.7\%) & $-$ & 
                        $-$ &  $-$  \\
                        & MIL~\cite{dong2017more}  & 92.81\% & 1.54\% & 7.81E7 (37.9\%) & 68.37\% & 2.96\% &  7.63E7 (39.3\%)  \\
                        & SFP~\cite{he2018soft}  & 93.35\% & 0.56\% & 5.94E7 (52.6\%) & 68.79\% & 2.61\% & 5.94E7 (52.6\%)  \\
                        & FPGM~\cite{he2019filter} & \textbf{93.49\%} & 0.42\% & 5.94E7 (52.6\%) & 69.66\% & 1.75\% & 5.94E7 (52.6\%)  \\
                        & GBN~\cite{you2019gate} & 93.43\% & \textbf{-}\textbf{0.33\%} & 5.09E7 (60.1\%) & - & 
                        - &  -  \\
                        & REPrune   & 91.85\% & 1.01\% & \textbf{4.35E7} \textbf{(65.9\%)} & \textbf{70.68\%} & \textbf{1.14\%} & \textbf{4.49E7} \textbf{(64.9\%)}  \\ \cline{1-8}
\multirow{5}{*}{110}    & MIL~\cite{dong2017more}  & 93.44\% & 0.19\% & 1.66E8 (34.2\%) & 70.78\% & 2.01\% &  1.73E8 (31.3\%)  \\
                        & PFEC~\cite{li2016pruning} & 93.30\% & 0.20\% & 1.55E8 (38.6\%) & $-$ & $-$ & $-$  \\
                        & SFP~\cite{he2018soft}  & 92.97\% & 0.70\% & \textbf{1.21E8} \textbf{(52.3\%)} & 71.28\% & 2.86\% & \textbf{1.21E8} \textbf{(52.3\%)}  \\
                        & FPGM~\cite{he2019filter} & 93.85\% & -0.16\% & \textbf{1.21E8} \textbf{(52.3\%)} & \textbf{72.55\%} & 1.59\% & \textbf{1.21E8} \textbf{(52.3\%)} \\
                        & REPrune & \textbf{94.09\%} & \textbf{-}\textbf{0.53\%} & 1.31E8 (49.0\%) & 71.76\% & \textbf{0.25\%} & 1.42E8 (44.7\%)  \\ \cline{1-8}
\end{tabular}
}
\end{center}
\caption{Comparison results of ResNet on CIFAR.}
\label{resnetcifar}
\end{table*}

{\bf ResNet on CIFAR}. We adopt soft fine-tuning \cite{he2018soft} to be generalized sufficiently and test REPrune on ResNet-20, 32, 56, and 110. As shown in Table~\ref{resnetcifar}, REPrune produces a comparable or better model for both CIFAR-10 and CIFAR-100. For example, FPGM \cite{he2019filter} on ResNet-20 shows 42.2\% FLOPs reduction with 1.11\% accuracy drop. On the contrary, REPrune achieves 46.0\% speedup with 1.06\% accuracy drop. Both PFEC \cite{li2016pruning} and MIL \cite{dong2017more} suffer from noticeable accuracy drop with at best 27.6\% and 37.9\% speedup. However, REPrune shows a comparable drop rate while achieving 65.9\% FLOPs reduction. For ResNet-110, REPrune has the accuracy gain by 0.53\% with 49.0\% reduced FLOPs, which outperforms FPGM \cite{he2019filter}. Regrading CIFAR-100, REPrune presents 1.14\% accuracy drops on ResNet-56, which achieves the best score. These results show that REPrune can achieve larger acceleration for comparable performance, or better performance for a comparable speedup, especially for hard-to-prune models like ResNet-56 and 110.

\begin{table}[!t]
\begin{center}
\scriptsize{
\begin{tabular}{c|c|cc|cc|c}
\hline
\multirow{2}{*}{Models} & \multicolumn{1}{c|}{\multirow{2}{*}{Methods}} & \multicolumn{2}{c|}{Top-1} & \multicolumn{2}{c|}{Top-5} & \multicolumn{1}{c}{\multirow{2}{*}{FLOPs}} \\ \cline{3-6}
                        & \multicolumn{1}{c|}{} & \multicolumn{1}{c|}{Pruned Acc.} & \multicolumn{1}{c|}{Acc. Drop} & \multicolumn{1}{c|}{Pruned Acc.} & \multicolumn{1}{c|}{Acc. Drop} & \multicolumn{1}{c}{} \\ \hline
\multirow{5}{*}{18}     & CCK~\cite{son2018clustering} & \textbf{69.90\%} & \textbf{-}\textbf{0.10\%} & \textbf{89.30\%} & \textbf{-}\textbf{0.20\%} & 1.44E9                         (21.2\%) \\
                        & MIL~\cite{dong2017more}  & 66.33\% & 3.65\% & 86.94\% & 2.29\% & 1.20E9 (34.2\%) \\
                        & SFP~\cite{he2018soft}  & 67.10\% & 3.18\% & 87.78\% & 1.85\% & \textbf{1.06E9} \textbf{(41.8\%)} \\
                        & FPGM~\cite{he2019filter} & 68.34\% & 1.94\% & 88.53\% & 1.10\% & \textbf{1.06E9} \textbf{(41.8\%)} \\
                        & REPrune & 68.09\% & 1.67\% & 87.99\% & 1.09\% & \textbf{1.06E9} \textbf{(41.8\%)} \\ \cline{1-7}
\multirow{7}{*}{50}     & ThiNet~\cite{luo2017thinet} & 72.04\% & 0.84\% & 90.67\% &
                        0.47\% & 2.61E9 (36.7\%) \\
                        & SFP~\cite{he2018soft} & 74.61\% & 1.54\% & 92.06\% & 0.81\% & 2.38E9 (41.8\%) \\
                        & FPGM~\cite{he2019filter} & \textbf{75.59\%} & \textbf{0.56\%} & \textbf{92.63\%} & \textbf{0.24\%} & 2.36E9 (42.2\%) \\
                        & NISP~\cite{yu2018nisp}& - & 0.89\% & $-$ & $-$ & 2.32E9 (44.0\%) \\
                        & Taylor~\cite{molchanov2019importance} & 74.50\% & 1.68\% & $-$ & $-$ & 2.25E9 (44.9\%) \\
                        & CP~\cite{he2017channel} & $-$ & $-$ & 90.80\% & 1.40\% & \textbf{2.06E9} \textbf{(50.0\%)} \\
                        & REPrune &  74.75\% & 1.40\% & 92.21\% & 0.66\% & 2.43E9 (41.2\%) \\ \cline{1-7}
\end{tabular}
}
\end{center}
\caption{Comparison results of ResNet on ImageNet.}
\label{resnetimagenet}
\end{table}

{\bf ResNet on ImageNet}. We evaluate REPrune on ResNet-18, 50. We only rely on mentioned selection criteria with  REPrune for ResNet-18, but, for ResNet-50, we do not prune identity mapping in shortcuts as well as last point-wise convolution to avoid dimension conflict in the bottleneck. Unlike ResNet on CIFAR, we do not fine-tune in a soft manner \cite{he2018soft}. As shown in Table~\ref{resnetimagenet}, REPrune outperforms or achieves comparable performance to prior works. For ResNet-18, REPrune achieves same acceleration level with \cite{he2018soft}, \cite{he2019filter}, but accuracy exceeds by 1.51\% and 0.27\%, respectively. For ResNet-50, REPrune has 1.40\% accuracy drops, which is better than \cite{he2018soft}, \cite{molchanov2019importance} with comparable speedup. Compared to soft norm-based pruning \cite{he2018soft}, considering filter distributions maintains the accuracy in both models. Also, for the geometric median \cite{he2019filter}, REPrune can select more meaningful filters in the smaller ResNet-18 model.

\begin{table}[!t]
\begin{center}
\scriptsize{
\begin{tabular}{c|cc|cc|c|c}
\hline
\multicolumn{1}{c|}{\multirow{2}{*}{Methods}} & \multicolumn{2}{c|}{Top-1} & \multicolumn{2}{c|}{Top-5} & \multicolumn{1}{c|}{\multirow{2}{*}{FLOPs}} & \multicolumn{1}{c}{\multirow{2}{*}{\#Params}} \\ \cline{2-5}
            & \multicolumn{1}{c|}{Pruned Acc.} & \multicolumn{1}{c|}{Acc. Drop} & \multicolumn{1}{c|}{Pruned Acc.} & \multicolumn{1}{c|}{Acc. Drop} & \multicolumn{1}{c|}{} & \multicolumn{1}{c}{} \\ \hline
ThiNet~\cite{luo2017thinet} & 69.80\% & \textbf{-1.46\%} & 88.44\% & \textbf{-1.09\%} & \textbf{4.85E9} \textbf{(69.0\%)} & 131.4E6 (5.0\%) \\ 
CP~\cite{he2017channel} & $-$ & $-$ & 89.90\% & 0.00\% & 7.83E9 (50.0\%) & $-$\\
$\ell_1$-norm & 72.15\% & 1.21\% & \textbf{90.84\%} & 0.68\% & 6.24E9 (60.1\%) & 91.1E6 (34.2\%) \\
REPrune & \textbf{72.25\%} & 1.11\% & 90.76\% & 0.76\% & 7.41E9 (52.7\%) & \textbf{78.6E6} \textbf{(43.1\%)} \\ \cline{1-7}
\end{tabular}
}
\end{center}
\caption{Comparison results of VGG-16 on ImageNet.}
\label{vggresults}
\end{table}

{\bf VGG on ImageNet}. 
REPrune obtains comparable or better performance. As shown in Table~\ref{vggresults}, CP~\cite{he2017channel} shows 89.90\% Top-5 accuracy with 50\% FLOPs reduction, and REPrune outperforms CP by 0.88\% Top-5 accuracy and 2.7\% FLOPs reduction. ThiNet~\cite{luo2017thinet} shows 69.0\% FLOPs reduction with performance gain, but it is limited to only 5\% of parameter optimization, which still shows memory hungry. For $\ell_1$-norm, it removes filters statically with the same pruning ratio determined by REPrune. Since REPrune remains filters dynamically, most filters in high-level layers, which determines the cost, are extremely pruned, but that in some layers are preserved. Thus, REPrune is beneficial for memory but, hard to reduce the FLOPs due to the computation by preserved filters.



\subsection{Test Accuracy Recovery Rate during Fine-tuning} \label{effectiveness}

\begin{figure*}[!t]
\centering{
\subfigure[] {\includegraphics[width=0.48 \columnwidth]{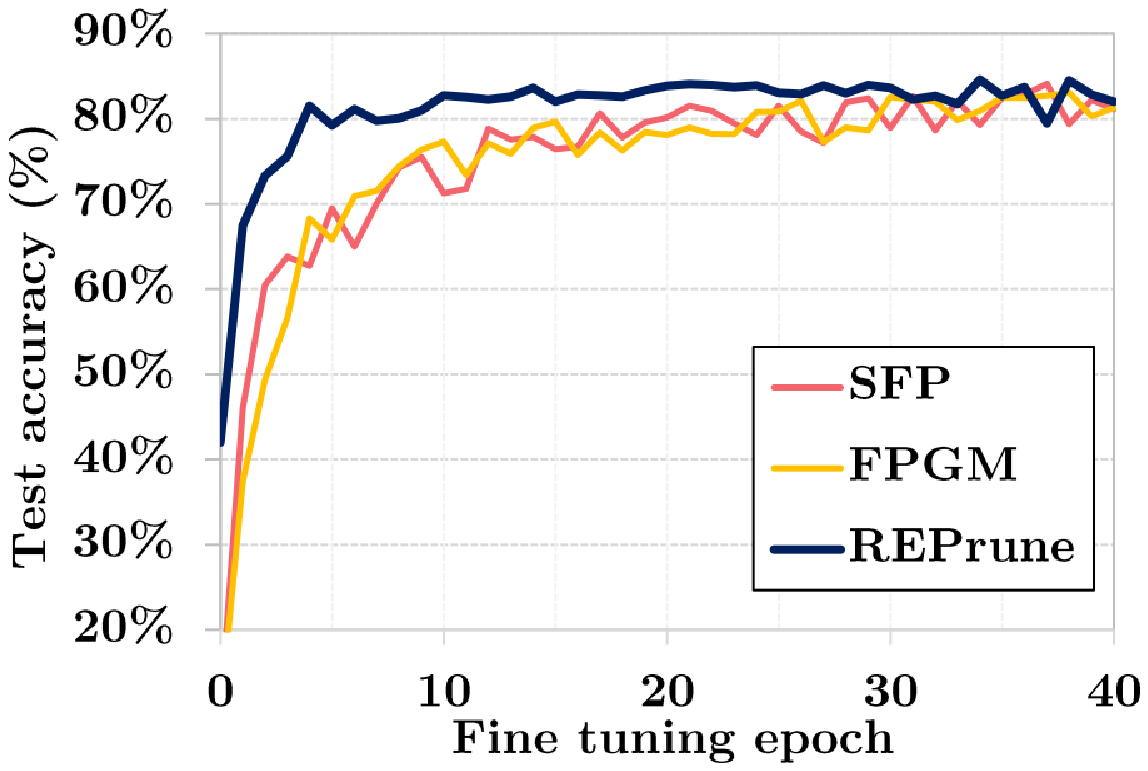}
\label{fig5_a}}
\subfigure[] {\includegraphics[width=0.48 \columnwidth]{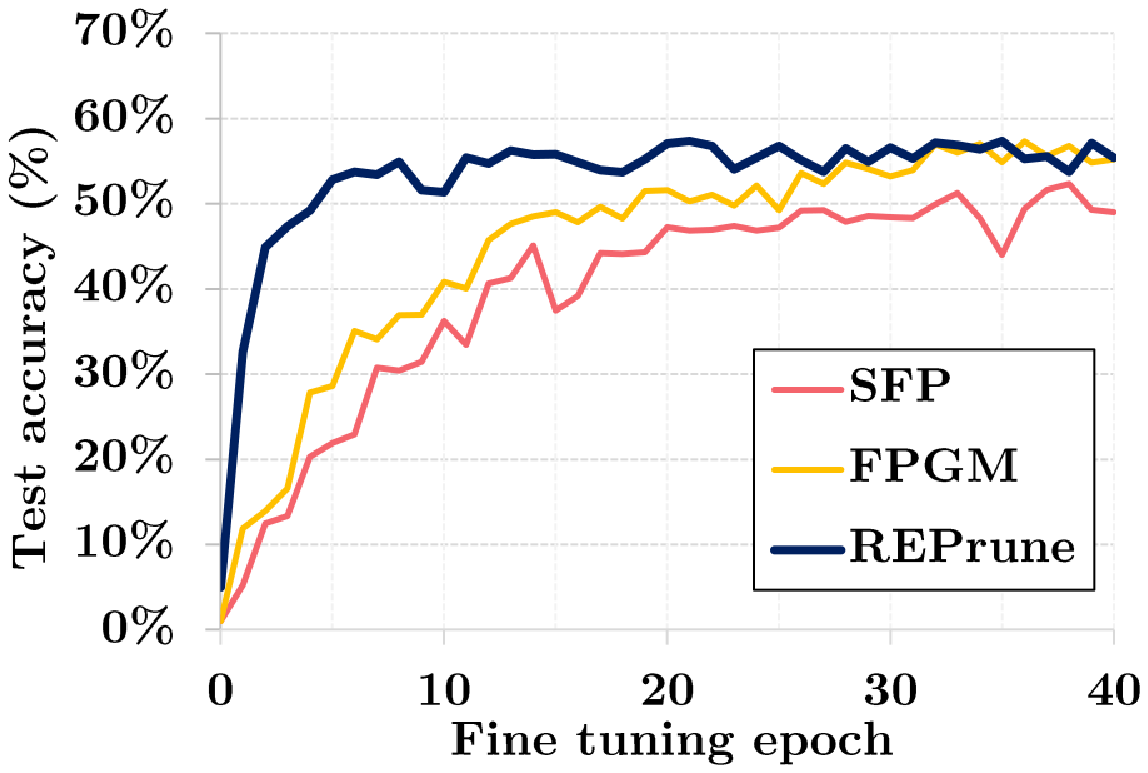}
\label{fig5_b}}
}
\caption{Validation accuracy recovery trends of VGG-16 on (a) CIFAR-10 and (b) CIFAR-100 during the first 40 epochs of fine-tuning.}
\label{fig5}
\end{figure*}

We also inspect the speed of validation accuracy recovery during the fine-tuning process. Our method is compared with the two $\ell_2$ norm-based filter pruning methods, SFP~\cite{he2018soft} and FPGM~\cite{he2019filter}. In this experiment, $\lambda$ of REPrune is set to 0.1. As shown in Fig.~\ref{fig5}, the VGG-16 model pruned by REPrune has a faster recovery rate than models using other techniques. This trend can be seen for both CIFAR-10 and CIFAR-100. In only 10 epochs of fine-tuning, REPrune reaches the accuracy level that other approaches can reach in 40 epochs. This rapid recovery of accuracy is especially useful when fine-tuning is limited due to time or computing resource constraints. Especially on CIFAR-10, validation accuracy of REPrune without fine-tuning exceeds that of other methods by more than 30\%.

\begin{figure*}[!t]
\centering{
\subfigure[] {\includegraphics[width=0.48 \columnwidth]{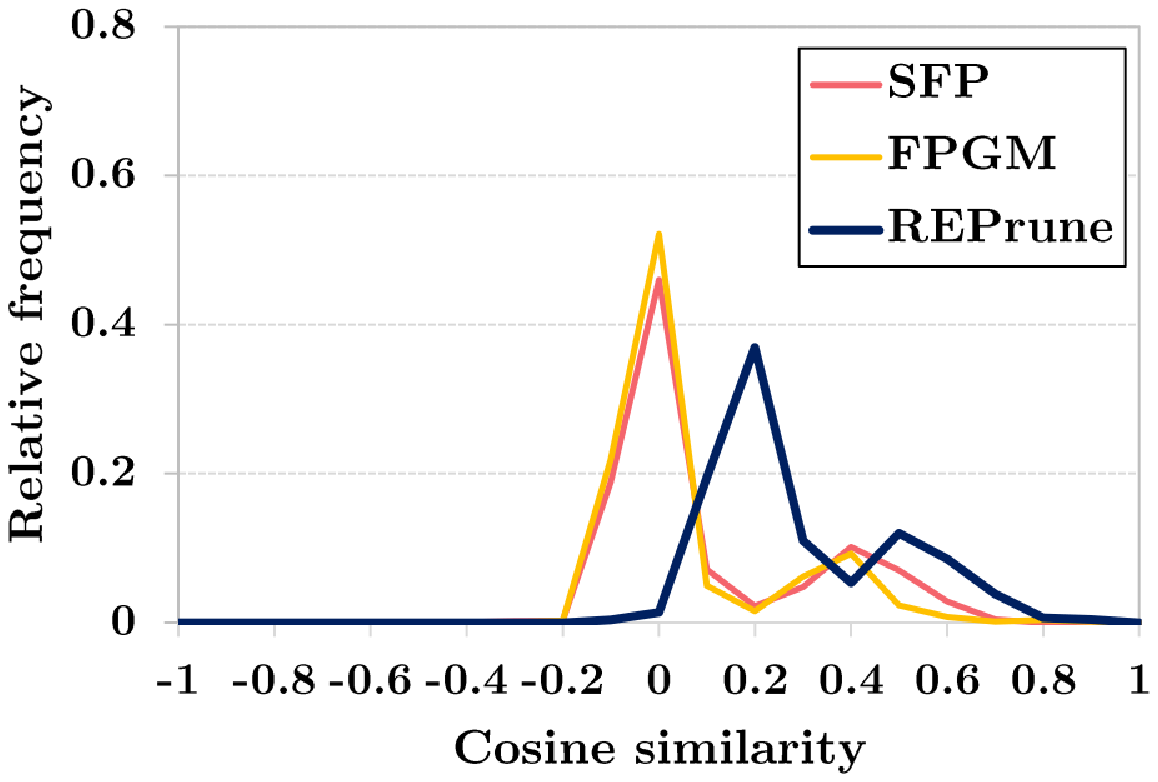}
\label{fig6_a}}
\subfigure[] {\includegraphics[width=0.48 \columnwidth]{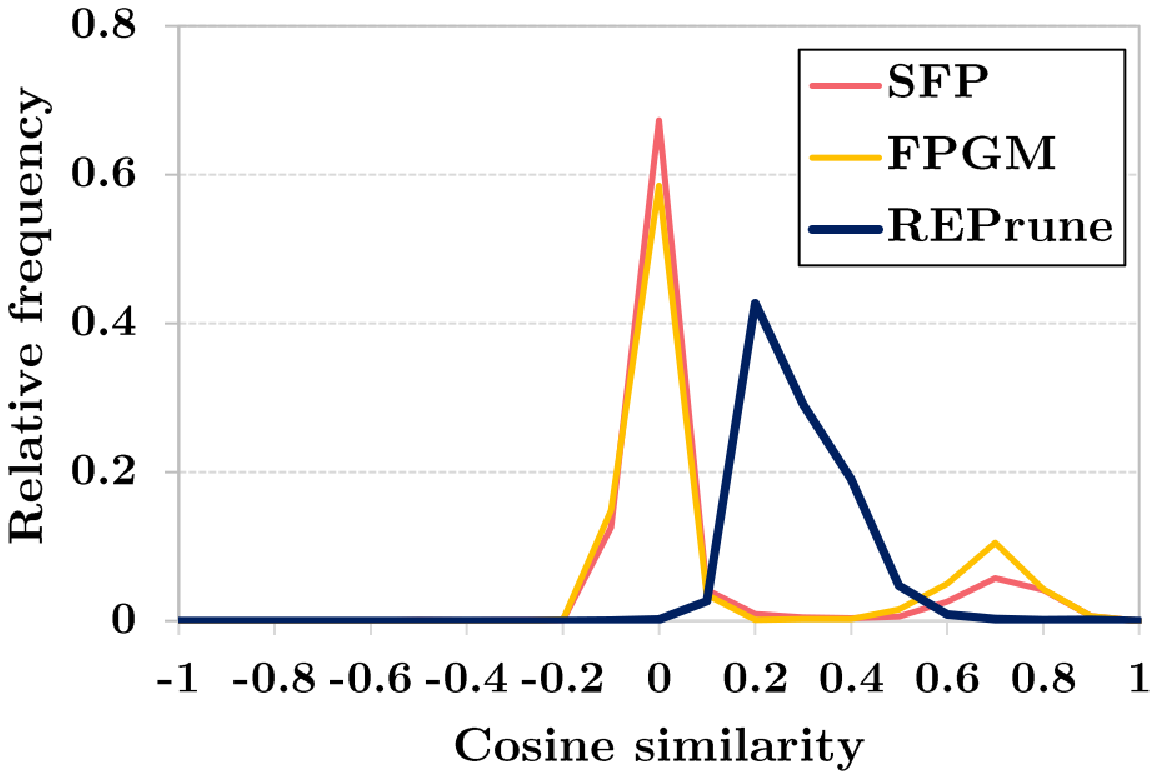}
\label{fig6_b}}
}
\caption{Histogram of cosine similarity between the filters before and after fine-tuning. All layers of VGG-16 on (a) CIFAR-10 and (b) CIFAR-100 are included.
}
\label{fig6}
\end{figure*}

\begin{figure*}[!t]
\centering{
\subfigure[VGG-16 on CIFAR-10] {\includegraphics[width=0.48 \columnwidth]{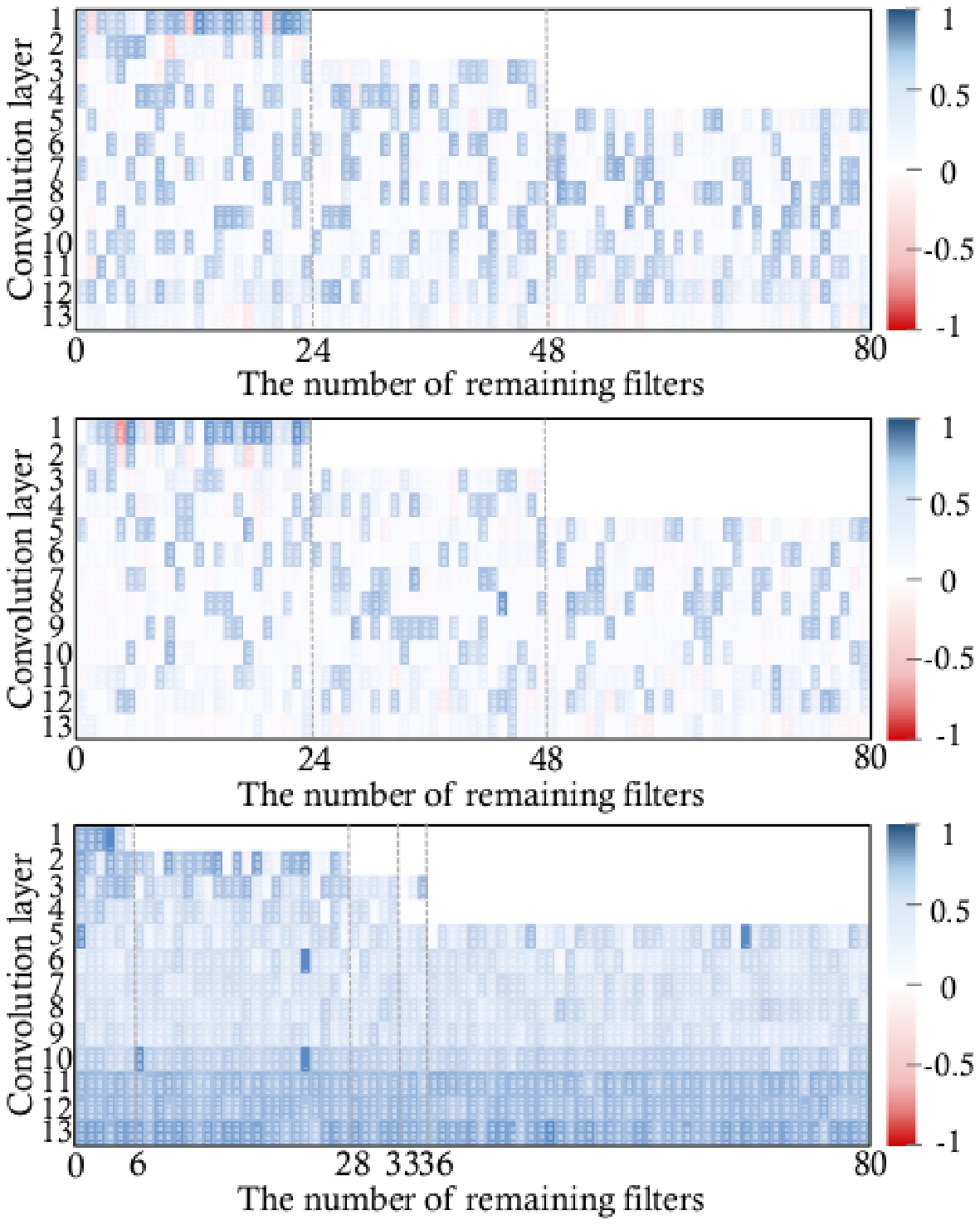}
\label{fig7_a}}
\subfigure[VGG-16 on CIFAR-100] {\includegraphics[width=0.48 \columnwidth]{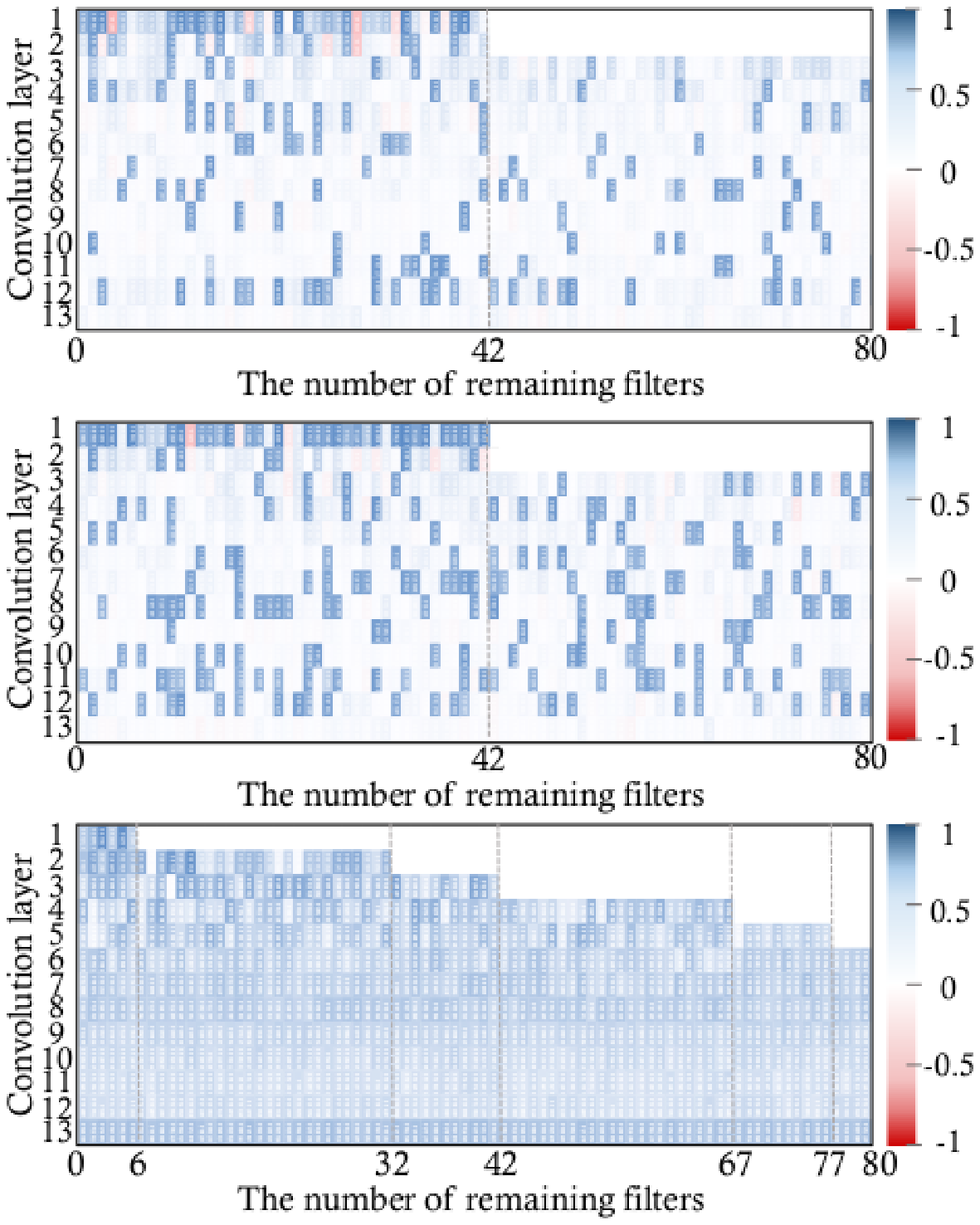}
\label{fig7_b}}
}
\caption{Visualization of cosine similarity between the filters before and after fine-tuning. Each of the top 3 row corresponds to the result of SFP, FPGM, and our method. If the color of each filter is closer to blue, the higher the similarity; if the color is closer to red, the lower the similarity. White indicates zero cosine similarity. Only the first 80 filters of each model are displayed since the pattern of the rest is similar.
}
\label{fig7}
\end{figure*}

REPrune enables the pruned model to recover its validation accuracy rapidly during fine-tuning. One possible explanation is due to its ability to maintain the original filter distribution. REPrune aims to leave filters that best represent the filter distribution of the full model. Therefore, remaining filters after the pruning do not move much during the fine-tuning process, which results in the rapidness of accuracy recovery. 

To verify this explanation, we analyze the cosine similarity of the remaining filters before and after fine-tuning. If the cosine similarity is 0, the two filters are completely orthogonal, and as the value approaches 1, the orientations of the two filters coincide. As shown in Fig.~\ref{fig6}, the average cosine similarity of REPrune is higher than those of other norm-based filter pruning methods. For REPrune on CIFAR-100, approximately 95\% of the cosine similarity of total filters is between 0.2 and 0.6. In the case of norm-based pruning, however, more than 70\% of the cosine similarity is between -0.1 and 0.1, e.g., 79.89\% for SFP and 73.45\% for FPGM. The result of CIFAR-10 also showed similar patterns. Thus, our method shifts the direction of the filter less than the norm-based pruning method.

This tendency is more evident in the cosine similarity heatmap, Fig.~\ref{fig7}. For SFP and FPGM, a large part of the heatmaps is white or light red, which indicates the low cosine similarity. However, the first convolution layer is the exception, as it extracts general features such as edges. For REPrune, however, its heatmap has a blue tone revealing that filters with high cosine similarity are evenly distributed across all layers. This tendency is seen in both CIFAR-10 and CIFAR-100, but due to the difference in the number of label classes, the color distribution is slightly different. This experiment confirms that REPrune actually selects high quality and informative filters which move less during fine-tuning.
\section{Conclusion and Future Work} \label{conclusion_futurework}
We demonstrate the limitations of previous norm-based pruning methods, addressing that norm alone cannot measure the importance of the convolution filter. In this paper, we propose REPrune that selects the filters that can best represent the distribution of the convolutional layer. Our method maintains the accuracy well under extreme pruning constraints. In addition, the filters selected by REPrune shift less during the fine-tuning process, allowing the test accuracy to converge rapidly. Considering its ability to represent the distribution of the filters, we plan to combine REPrune with knowledge distillation for the pruned model to be more generalized.

\clearpage
\bibliographystyle{splncs04}
\bibliography{egbib}
\end{document}